%% file: main.tex
\documentclass[10pt,twocolumn,letterpaper]{article}

\usepackage[pagenumbers]{cvpr} 

\usepackage{diagbox}
\usepackage{svg}
\usepackage{comment}
\usepackage{dsfont}

\usepackage{amsmath,amssymb,amsfonts}
\usepackage{graphicx}
\usepackage{textcomp}
\usepackage{xcolor}

\usepackage{algorithm}
\usepackage{algpseudocode}

\usepackage{url}
\usepackage{subcaption}
\usepackage{booktabs}
\usepackage[edges]{forest}
\usepackage{xcolor}
\usepackage{diagbox}
\usepackage{tikz}

\input{preamble}

\definecolor{cvprblue}{rgb}{0.21,0.49,0.74}
\usepackage[pagebackref,breaklinks,colorlinks,citecolor=cvprblue]{hyperref}

\def\paperID{6WD28XudKk} 
\def\confName{3DV\xspace}
\def\confYear{2027\xspace}

\title{A Real-World CAD Segmentation Benchmark: Comparing Rule-Based and Learning-Based Approaches through Hole Detection}

\author{Kai Schütte\\
Volkswagen AG\\
Universität Osnabrück\\
{\tt\small kai.schuette@volkswagen.de}
\and
Edgar Heinert\\
Universität Osnabrück\\
{\tt\small edgar.heinert@uni-osnabrueck.de}
\and
Matthias Rottmann\\
Universität Osnabrück\\
{\tt\small matthias.rottmann@uni-osnabrueck.de}
}

\author{
    Kai Schütte$^{1,2}$ \quad Edgar Heinert$^{2}$ \quad Matthias Rottmann$^{2}$ \\
    {\normalsize $^1$Volkswagen AG},  
    {\normalsize $^2$Universität Osnabrück} \\ 
    {\tt\small kai.schuette@volkswagen.de, \{edgar.heinert, matthias.rottmann\}@uni-osnabrueck.de}
}

\begin{document}
\maketitle
\input{sec/0_abstract}    
\input{sec/1_introduction}
\input{sec/2_related_work}
\input{sec/3_methodology}
\input{sec/4_annotation}

\input{sec/5_experiments}
\input{sec/6_conclusion}

{
    \small
    \bibliographystyle{ieeenat_fullname}
    \bibliography{main}
}

\input{sec/7_appendix}


\end{document}

%% file: preamble.tex
\usepackage[dvipsnames]{xcolor}


%% file: sec/0_abstract.tex
\begin{abstract}
CAD segmentation datasets are predominantly synthetic or automatically labelled, leaving it unclear how well existing methods transfer to real-world engineering tasks. Fastener-relevant hole detection represents one such task, requiring human annotation and exhibiting greater geometric complexity than commonly used benchmarks. We address this gap through three main contributions. First, we develop two task-specific rule-based methods, one graph-wire-search and one multi-view based, and complement them with a human-annotated benchmark dataset derived from engineering CAD data that also enables training of UV-Net and AAGNet, representative learning-based approaches for B-Rep segmentation. Second, we conduct an extensive experimental study including parameter optimisation, evaluation on our benchmark dataset, cross-domain generalisation on a synthetic dataset, and an assessment of the feasibility of the approaches at varying dataset sizes and IoU thresholds.
Third, we provide evidence-based insights into the strengths and limitations of different approaches: rule-based methods generalise well without training data, with the graph-wire-search method favouring recall and the multi-view method favouring precision, whereas learning-based methods achieve strong performance with sufficient annotations but exhibit reduced robustness under domain shift. These findings highlight trade-offs between data requirements, and robustness, and provide guidance not only for hole detection, but more broadly for real-world CAD segmentation tasks.
\end{abstract}


%% file: sec/1_introduction.tex
\section{Introduction}
\label{sec:intro}

Recent advances in CAD segmentation and machining feature recognition have been enabled by the availability of large benchmark datasets. However, most existing datasets are synthetic or derived from publicly shared CAD models, and their labels are typically generated automatically from CAD construction histories \cite{lambourne2021brepnet, hierarchicalcadnet2022, ZHANG2024102318}. While this allows for large-scale training and evaluation, it is unclear how well the resulting methods transfer to real-world engineering tasks, where labels are often application-specific, manually created, and subject to ambiguity.
\begin{figure}[t]
    \centering
    \begin{subfigure}[b]{0.45\linewidth}
        \centering
        \includegraphics[width=\linewidth]{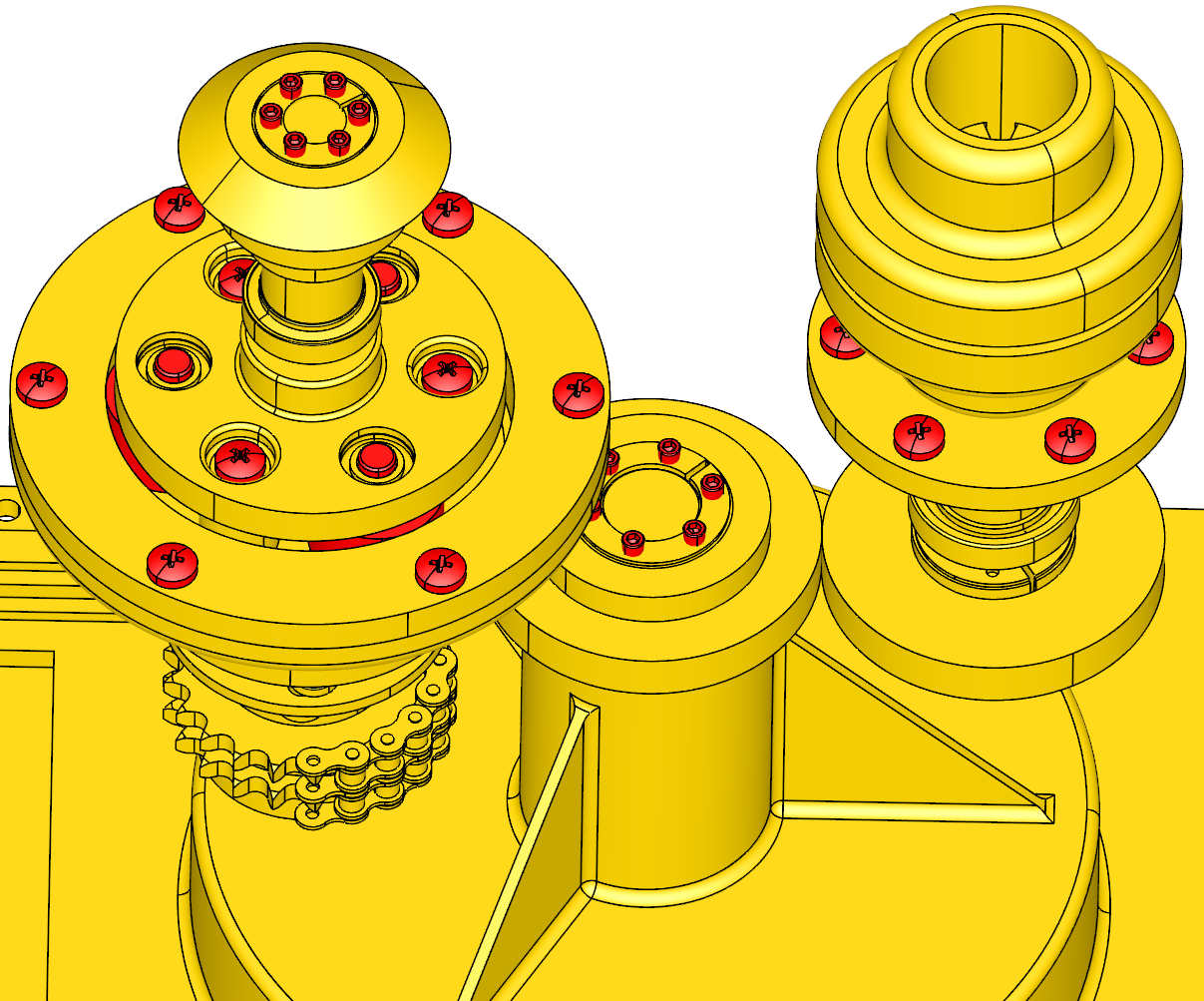}
        \caption{Screws on an assembly highlighted in red}
        \label{fig:assembly_fasteners}
    \end{subfigure}
    \hfill
    \begin{subfigure}[b]{0.45\linewidth}
        \centering
        \includegraphics[width=\linewidth]{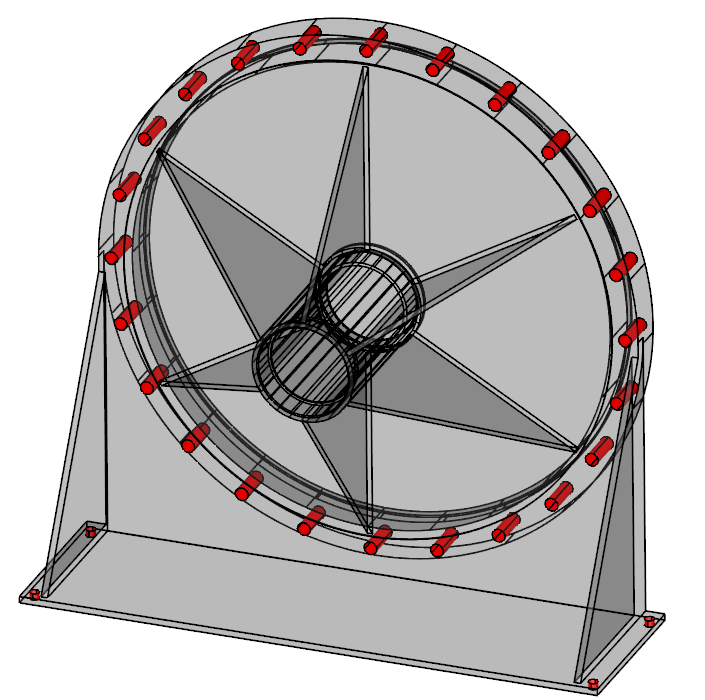}
        \caption{Part with fastener holes highlighted in red}
        \label{fig:part_with_holes}
    \end{subfigure}
    \caption{Overview of screws and fastener holes}
    \label{fig:fasteners_and_holes}
\end{figure}
%
%
Many engineering workflows require semantic annotations that cannot be reliably recovered from modelling operations alone \cite{heidariGeometricDeepLearning2024}. These include fastener-relevant holes \cite{https://doi.org/10.48550/arxiv.2111.12772}, surfaces relevant to defeaturing before mesh generation \cite{owenSurveyAIMethods2025}, and other functionally defined surface regions. Manually creating such annotations is costly and often results in limited training data and a lack of representative benchmarks for evaluating learning-based methods.

To investigate these challenges, we consider fastener-relevant hole detection as a representative real-world CAD segmentation task.
%
Holes frequently occur in CAD models, either as intentional design features \cite{https://doi.org/10.48550/arxiv.2111.12772} or as artefacts introduced during modelling and format conversion \cite{CAMBA201618}. Their reliable identification is challenging due to the wide variety of hole geometries \cite{GHADAI2018263,Lai2020} and is particularly important for assembly modelling, where hole information reflects fastener compatibility and assembly feasibility (\eg in car manufacturing) \cite{https://doi.org/10.48550/arxiv.2111.12772,SolidWorks_SmartFasteners_2025}. 
Modern CAD systems exploit such information for automated fastener placement and assembly assistance, but existing hole-recognition tools are typically limited to comparatively simple cylindrical features \cite{AutodeskFusion360_HoleRecognition, Suradkar2018_RecognizingHoles_SE}. As a result, many fastener-relevant hole types remain difficult to detect automatically, despite occurring frequently in engineering assemblies, including geometries associated with screws, rivets, clips, and bolts (Figure~\ref{fig:fasteners_and_holes}). Their reliable identification is essential for automated assembly workflows, especially in industries with large heterogeneous models, such as aerospace or automotive.
Prior work on hole detection has mainly focussed on manufacturability analysis \cite{GHADAI2018263,PRABHU199925,LYNN20161183}, process planning \cite{SUNIL2008502, SUNIL2010686}, meshing \cite{Lai2019,Lai2020,moriya2023graph} and additive manufacturing \cite{DING2016139}. Hole information has also been linked to higher-level applications like CAD search \cite{MILES20243239,Lee2022} or assembly guidance \cite{songAutomatedGuidancePeginhole2014}.
Existing strategies fall into two families. Rule-based methods, which use geometric heuristics \cite{Lai2019,Lai2020,Kukreja2021,Ibrahim2021, tan2007development,reddy2014design} and are efficient and interpretable, but struggle with non-standard shapes and learning-based methods, which \cite{9577057,ZHANG201812,lambourne2021brepnet,moriya2023graph} offer flexibility, but require extensive labelled data and are sensitive to domain shift \cite{ZHANG2024102318}.However, existing evaluations are predominantly conducted on synthetic or automatically labelled datasets, leaving open how these approaches perform on manually annotated engineering CAD data. This motivates a systematic study of real-world CAD segmentation through the task of fastener-relevant hole detection.
%
To close this gap, we systematically compare rule-based and learning-based approaches for fastener-relevant hole detection. We introduce a labelled real-world dataset, adapt UV-Net to the task, and develop two novel rule-based algorithms. This enables the first systematic comparison of the two paradigms in terms of accuracy, robustness, interpretability, and sensitivity to domain shift. Our study highlights the respective strengths and weaknesses of each approach and clarifies their suitability for automated fastener placement in heterogeneous CAD environments.

\textbf{Our contributions are:}
\begin{itemize}
    \item We develop two rule-based methods. One graph-based and one computer-vision-based method and complement them with the annotation and release of a manually annotated dataset.
    \item We conduct an extensive experimental study, including parameter optimisation for each method, evaluation on our benchmark dataset, an analysis of  generalisation on a subset of the MFCAD++ dataset \cite{hierarchicalcadnet2022}, an assessment of the impact of dataset size, and an investigation of IoU thresholds on detection quality.
    \item We provide evidence-based insights into the strengths and limitations of different approaches: rule-based methods deliver stable cross-domain performance, with the graph-wire-search method favouring recall and the multi-view method favouring precision, while UV-Net achieves strong accuracy with sufficient data but struggles to generalise across heterogeneous datasets.
\end{itemize}

%% file: sec/2_related_work.tex
\section{Related Work}
\label{sec:related_work}
Research on hole detection and feature recognition in CAD models spans a wide range of data representations and techniques, including Boundary Representation (BRep), meshes, point clouds, and voxel grids. This section first reviews existing CAD segmentation datasets before discussing rule-based and learning-based approaches.

\subsection{CAD Segmentation Datasets}
The development of learning-based CAD segmentation methods has been enabled by the availability of benchmark datasets for machining feature recognition and semantic segmentation. Existing datasets are commonly derived either from synthetic CAD models with automatically generated labels or from publicly shared CAD repositories. Such datasets facilitate large-scale training and evaluation, but may not fully reflect the complexity and ambiguity encountered in engineering applications.
Prominent examples include MFCAD \cite{caoGraphRepresentation3D2020}, MFCAD++ \cite{hierarchicalcadnet2022}, Fusion 360 Gallery \cite{lambourne2021brepnet}, and CADSynth \cite{ZHANG2024102318}. These datasets derive labels directly from CAD construction histories, enabling scalable annotation while avoiding many challenges associated with task-specific human annotations. Consequently, the extent to which methods evaluated on these benchmarks generalise to real-world engineering tasks remains unclear.

\begin{figure}[!htbp] 
\centering 
\includegraphics[width=\columnwidth]{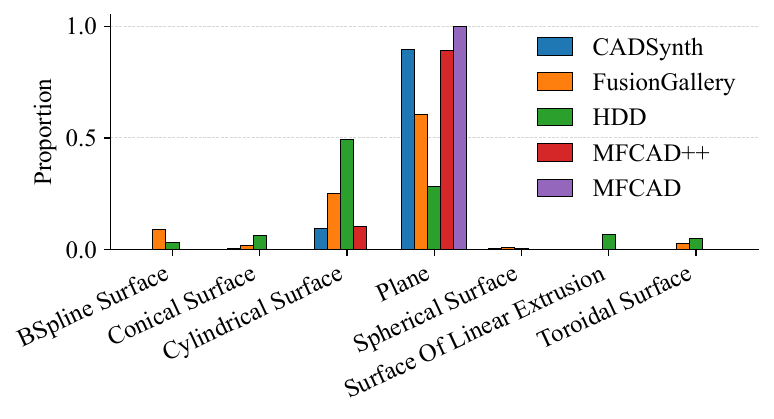}
\vspace{-2mm} 
\includegraphics[width=\columnwidth]{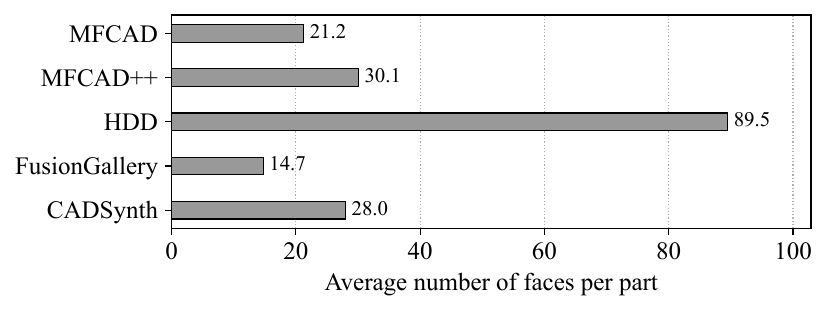}
\caption{Comparison of publicly available CAD segmentation datasets. Top: face-type distribution. Bottom: average number of faces per part.} 
\label{fig:dataset_stats}
\end{figure}
Beyond annotation strategy, the geometric characteristics of these datasets also differ substantially from those of our benchmark. As shown in Figure~\ref{fig:dataset_stats}, common CAD segmentation benchmarks are largely dominated by planar surfaces and contain considerably fewer faces per part than our benchmark. Together, these observations indicate that manually annotated engineering CAD data introduces both greater geometric diversity and increased structural complexity than commonly used benchmark datasets.
%
%

%
\subsection{Rule-Based Approaches}
Rule-based methods rely on explicit geometric or topological rules. Several works detect holes using convex loops or closed wires \cite{Lai2019,Lai2020}, or through cylindrical pattern matching \cite{Kukreja2021}. Other approaches restrict themselves to narrowly defined geometries, such as cross-holes in hollow cylinders \cite{Ibrahim2021}, planar cylindrical features \cite{tan2007development}, or stepped and circular holes \cite{reddy2014design}. While such methods are efficient and interpretable, they struggle with non-standard geometries or with variations introduced by different modelling conventions across CAD software. The rule-based methods investigated in this work follow a similar philosophy by prioritising topological consistency over strict geometric assumptions, with the goal of improving robustness across heterogeneous engineering datasets.
\subsection{Learning-Based Approaches}
Learning-based methods have become a popular approach for CAD segmentation. UV-Net \cite{9577057} uses UV parameterisations of faces for feature classification, but its embeddings are not inherently rotation-invariant, which reduces robustness in real-world scenarios. Other works extend graph-based learning in different ways. Moriya \etal\ \cite{moriya2023graph} model BRep structures with GNNs, achieving strong overall accuracy but reporting weaker recall for holes, while Zhang \etal\ \cite{ZHANG2024102318} incorporate transformers and domain adaptation strategies that are mostly validated on synthetic data.
Voxel-based methods such as FeatureNet \cite{ZHANG201812} use 3D CNNs on synthetic cube datasets, which limits geometric variation. Reported metrics cover only accuracy, making recall unclear.
In subsequent studies \cite{9577057}, UV-Net has been shown to outperform FeatureNet under more comprehensive evaluation. Ghadai \etal\ \cite{GHADAI2018263} also apply 3D CNNs to voxel data, targeting manufacturability analysis. Their approach highlights regions related to holes but does not explicitly identify hole faces or their geometry for part connectivity.
BRepNet \cite{lambourne2021brepnet} encodes faces, edges, and co-edges directly while omitting coordinates to ensure invariance, yet it struggles on rare or hole-specific classes. Hierarchical models, such as Miles \etal\ \cite{milesApproachingSTEPFile2023}, leverage STEP file trees and are better suited to global part properties than to fine-grained hole detection.
Overall, learning-based methods provide greater flexibility than rule-based approaches but depend on representative annotations and sufficient training data. 
Their applicability to real-world engineering tasks remains unclear, due to evaluations being predominantly conducted on synthetic or automatically labelled datasets.


%% file: sec/3_methodology.tex
\section{Methods}
\label{sec:methodology}
We compare three approaches for fastener-relevant hole detection in CAD models: two task-specific rule-based methods and two learning-based baselines. Together, they enable an analysis of the trade-offs between handcrafted and data-driven approaches on a real-world segmentation task. The graph-wire-search (GWS) method relies on topology, the multiview rendering method (MVR) exploits rendered views, and UV-Net operates directly on surface parametrisations. Despite these differences, they share a common representation, introduced below before each method is described in detail.
\subsection{Graph Representation}
\label{ssec:graph_representation}
All three methods rely on a graph-based representation of CAD models. 
Let the set of faces be
\begin{equation}
\mathcal{F} = \{F_1, F_2, \dots\}, 
\end{equation}
and the set of edges be
\begin{equation}
\mathcal{E} = \{e_1, e_2, \dots\}.
\end{equation}
Each face $F_i$ is bounded by a subset of edges $E_{F_i} \subseteq \mathcal{E}$.  
Two faces $F_i, F_j \in \mathcal{F}$ are considered adjacent if they share at least one edge, which is expressed as
\begin{equation}
N(F_i, F_j) = 
\begin{cases}
1, & E_{F_i} \cap E_{F_j} \neq \emptyset, \\
0, & \text{otherwise}.
\end{cases}
\end{equation}
The neighbourhood relation is then defined as
\begin{equation}
\mathcal{N} = \{ (F_i, F_j) \in \mathcal{F} \times \mathcal{F} \mid N(F_i, F_j) = 1 \}.
\end{equation}
Finally, the \emph{face adjacency graph} is
\begin{equation}
G_F = (\mathcal{F}, \mathcal{N}),
\end{equation}
where nodes correspond to faces and edges represent adjacency. Such graph-based representations of CAD models are well established in the literature~\cite{10.1145/325165.325218,Ansaldi1985An}, and capture only topological information, ensuring invariance to translation, rotation, and scaling.
\subsection{Graph-Wire-Search Hole Detection}
The graph-wire-search method detects holes primarily from topological information in $G_F$, but includes a hollowness validation that uses local face normals. 
This provides invariance to translation, rotation, and scaling, while exploiting minimal geometric cues for verification. 
The procedure consists of four main steps: 1) identifying candidate wires, meaning closed edge loops which may correspond to hole openings, 2) validating hollowness, 3) expanding hole faces, and 4) classifying through- vs. blind-holes.
\paragraph{Closed wire extraction:}
Candidate hole openings are represented as closed wires, \ie, cycles of connected edges that stem from one or more adjacent faces. 
For a set of faces $\mathcal{F}^* \subseteq \mathcal{F}$, with $n_{\mathcal{F}^*} = \lvert \mathcal{F}^* \rvert$ the enclosing edges are
\begin{equation}
\mathcal{E}_{\mathcal{F}^*} = \bigcup_{F \in \mathcal{F}^*} \mathcal{E}_F .
\label{eq:edges_faces}
\end{equation}
Edges shared by two or more faces are removed, as they cannot bound a hollow region,
\begin{equation}
W_{\mathcal{F}^*} = \mathcal{E}_{\mathcal{F}^*} \setminus \left\{ e \mid \#\{ F^* \in \mathcal{F}^* \mid e \in \mathcal{E}_F \} > 1 \right\}.
\label{eq:wire_edges}
\end{equation}
The undirected graph induced by the remaining edges is
\begin{equation}
H_{\mathcal{F}^*} = (V^*, W_{\mathcal{F}^*}), \quad 
V^* = \{ v \mid v \in e, \; e \in W_{\mathcal{F}^*} \}.
\label{eq:wire_graph}
\end{equation}
Since all shared edges were removed, $H_{\mathcal{F}^*}$ typically splits into disconnected subgraphs, each representing a closed wire,
\begin{equation}
C(H_{\mathcal{F}^*}) = \left( C_1, C_2, \dots \right).
\label{eq:connected_components}
\end{equation}
If $|C(H_{\mathcal{F}^*})|=1$ the wire is rejected, as it only encloses the faces $F*$.
To keep the procedure efficient, the number of faces in $\mathcal{F}^*$ is limited by a parameter $n_{\mathrm{max}}$.
\paragraph{Hollowness validation:}
Each wire is validated by testing whether the region enclosed by adjacent faces is hollow. Faces are tessellated into triangles $\mathcal{T}$, and for each triangle $t$, centroid $\vec{c}_t$, normal $\vec{n}_t$ (pointing outward, into the cavity for holes), and area $A_t$ are computed. Normals are offset from centroids to form two point sets,  
\begin{equation}\label{eq:hollow_offset} 
\mathcal{P}^+ = \{\vec{c}_t + \vec{n}_t \}, \quad \mathcal{P}^- = \{\vec{c}_t - \vec{n}_t \}, 
\end{equation}
and the distributions of $\mathcal{P}^+$ and $\mathcal{P}^-$ are compared via the weighted variance  
\begin{equation}\label{eq:hollowness_variance} 
\mathrm{Var}_w(\mathcal{P}) = \frac{\sum_{t} A_t \, \|\vec{p}_t - \mu\|^2}{\sum_t A_t}, \quad \mu = \frac{\sum_t A_t \vec{p}_t}{\sum_t A_t}, 
\end{equation}  
where triangle areas serve as weights so that large surface patches dominate over small tessellation artefacts. If 
\begin{equation}\label{eq:hollowness_test}
\mathrm{Var}_w(\mathcal{P}^-) > \lambda \, \mathrm{Var}_w(\mathcal{P}^+),
\end{equation}
the region is considered hollow, since outward- and inward-offset normals spread differently when a cavity is present. The factor $\lambda > 1$ acts as a safeguard, as it prevents nearly planar regions from being misclassified as hollow, when the variance difference between $\mathcal{P}^+$ and $\mathcal{P}^-$ is small. It should be noted that choosing a large value for $\lambda$ may lead to small holes being rejected if the spread of their face normals is too small. In practice this did not occur, since all face normals were normalised. If such issues arise, scaling the normals by an appropriate factor would mitigate the problem.
\paragraph{Hole face expansion:}
Validated wires are iteratively expanded into full hole faces by following adjacency in $G_F$. The faces corresponding to the wire $F^*$ are excluded to ensure only faces in the hole direction are added. For a wire $C_j \in C$, the seed set of hole faces is
\begin{equation}
\mathcal{F}_{h}^{(0)} = \{ F \in \mathcal{F} \mid \exists \, e \in E_F: \, e \in C_j \} \setminus \mathcal{F}^* ,
\end{equation}
which collects all faces incident to the wire edges but not part of $F^*$. Expansion then proceeds by adding faces that are connected through at least one edge outside the cycle $C$, expressed as  
\begin{equation}
\mathcal{F}_{h}^{(i)} = \{\, F \in \mathcal{F}_{h}^{(i-1)} \mid \exists\, e \in E_F \setminus C \,\}.
\end{equation}
The process stops when no new faces can be added, and the final set $\mathcal{F}_h$ contains all faces belonging to the hole.
\paragraph{Through-hole check:}
Finally, the hole is classified as through or blind by testing whether any hole face connects to a non-hole face. 
We define the neighbouring face set as
\begin{equation}
\mathcal{F}_{\mathcal{N}_h} = \{ F_j \in \mathcal{F} \mid (F_i, F_j) \in \mathcal{N}, \; F_i \in \mathcal{F}_h \}.
\end{equation}
External faces are then obtained as
\begin{equation}
\mathcal{F}_{\mathrm{ext}} = \mathcal{F}_{\mathcal{N}_h} \setminus (\mathcal{F}_h \cup \mathcal{F}_e).
\end{equation}
If $\mathcal{F}_{\mathrm{ext}} \neq \emptyset$, the hole is marked as a through-hole, otherwise it is blind. 
As the benchmark labels only include through-holes, blind-holes are discarded.
Finally, the hollowness check is applied again to the complete set of detected hole faces $\mathcal{F}_h$ to ensure that the region is indeed hollow.
\subsection{Multiview Rendering (MVR) Method}
\label{sec:cv_method}
The second rule-based approach leverages rendered images to propose hole faces from visual contours, then validates them with BRep topology and a hollowness test.
\paragraph{Rendering and camera sampling} 
Each CAD part is tessellated, and all polygons belonging to a B-Rep face are assigned a unique colour for later identification. To render $n_\mathrm{img}$ images, the same number of camera positions is sampled uniformly on a sphere using the spherical Fibonacci method \cite{keinertSphericalFibonacciMapping2015}. The working distance $WD$ is set so that the part fits within the linear field of view defined by the bounding-box diagonal $d_{\mathrm{BB}}$. With $f$ denoting the camera focal length and $H$ the sensor height, the relation
\begin{equation}
    WD = \frac{f \cdot d_{\mathrm{BB}}}{H}
\end{equation}
ensures full visibility of the object. The derivation of this expression is provided in Appendix B.
To ensure that the smallest relevant hole of diameter $d_{min}$ still covers at least one pixel when observed at the maximum object distance $d_{o,\mathrm{max}}$, we set
\begin{equation}
d_{o,\mathrm{max}} = WD + \tfrac{1}{2}BB_{diag},
\end{equation}
which leads to the requirement
\begin{equation}
N \geq \left\lceil \frac{d_{o,\mathrm{max}} \cdot H}{f \cdot d_{min}} \right\rceil ,
\end{equation}
where $N$ denotes the number of pixels per image dimension.
\paragraph{Contour extraction and view consistency:}
For each image $I_i$, we detect edge pixels and read their face colours to obtain candidate faces $\mathcal{F}^{(i)}_{\mathrm{cand}}$. To suppress invalid contours, a face $F$ is kept only if it appears from sufficiently different viewpoints, which we enforce by requiring
\begin{equation}
\angle(\mathbf{v}_i,\mathbf{v}_j)=\arccos\!\left(\frac{\mathbf{v}_i^\top \mathbf{v}_j}{\|\mathbf{v}_i\|\,\|\mathbf{v}_j\|}\right)\ge \tau_\theta,
\end{equation}
where $\mathbf{v}_i$ is the camera viewing direction for image $I_i$ (pointing from the camera position to the model centre) and $\tau_\theta$ is an angle threshold~\cite{antonElementaryLinearAlgebra2014}.  
The retained set is then
\begin{equation}
\mathcal{F}_{\mathrm{cand}}=\{F \mid F\in \mathcal{F}^{(i)}_{\mathrm{cand}}\cap \mathcal{F}^{(j)}_{\mathrm{cand}},\, \angle(\mathbf{v}_i,\mathbf{v}_j)\ge \tau_\theta\}.
\end{equation}
\paragraph{Topology-aware clustering.}
The candidate faces $\mathcal{F}_{\mathrm{cand}}$ are first grouped into clusters $\mathcal{F}_k$ by traversing adjacency in the face adjacency graph $G_F$. In subsequent steps, each cluster is examined to determine whether it forms a valid hole.
For each cluster $\mathcal{F}_k$ we apply \crefrange{eq:edges_faces}{eq:connected_components} with $\mathcal{F}^*$ replaced by $\mathcal{F}_k$, which yields $C(H_{\mathcal{F}_k})$, representing wires.
If $|C(H_{\mathcal{F}_k})| = 2$, these two components are interpreted as the entrance and exit of a through-hole, otherwise the cluster is discarded.
A final hollowness check, using the variance criterion in Eq.~\eqref{eq:hollowness_variance}, verifies that the cluster encloses a cavity, with the offset construction of Eq.~\eqref{eq:hollow_offset} providing the test points.
This method benefits from multi-view redundancy and is agnostic to face parametrisations, but it is sensitive to rendering coverage and resolution. The structural checks ensure precision, while view consistency and hollowness control false positives from accidental contours.
\subsection{Learning-Based Methods}
\label{sec:uvnet_method}   
    %
UV-Net \cite{9577057} is a widely used point of reference for learning directly from boundary-representation (B-rep) geometry \cite{ZHANG2024102318, hierarchicalcadnet2022}. It represents each face by sampling its native parametric (u,v) domain into a regular 2D grid containing geometric channels such as position, normals, and trimming information, while each topological edge is represented by a sampled 1D grid along its analytic curve. Separate CNNs encode faces and edges before two message-passing layers aggregate information from a two-hop neighbourhood in the face-adjacency graph.
%
%
AAGNet \cite{WU2024102661} also operates directly on B-rep geometry using a geometric Attributed Adjacency Graph. It combines sampled face geometry with explicit face and edge attributes and in its default configuration, applies three message-passing layers, resulting in a three-hop local receptive field. Both models additionally concatenate a globally pooled graph representation with each face embedding before predicting per-face classes. Compared with UV-Net, AAGNet therefore incorporates explicit B-rep attributes and aggregates local information over a larger topological neighbourhood. Together, UV-Net and AAGNet provide representative learning-based approaches.

%% file: sec/4_annotation.tex
\section{Real-World CAD Segmentation Benchmark}\label{sec:benchmark}
To evaluate CAD segmentation methods under realistic engineering conditions, we created a manually annotated benchmark derived from assemblies in the 3D Assembly Repository \cite{3dassemblyrepo,LUPINETTI2019103111}. Unlike existing benchmark datasets, these assemblies provide greater geometric diversity and functional context, which is used to reduce bias in the annotation process. The following subsections describe the benchmark design, part selection and annotation policy.
\subsection{Benchmark Design and Annotation}
To address these limitations, we created a manually annotated benchmark derived from assemblies in the 3D Assembly Repository \cite{3dassemblyrepo,LUPINETTI2019103111}. Unlike isolated part models, assemblies provide functional context, enabling a binary segmentation of B-Rep faces into \emph{hole} and \emph{non-hole} classes.
The dataset contains a variety of mechanical assemblies, often joined via standard fasteners such as screws, enabling annotation based on functional context rather than purely geometric criteria. The benchmark is not without limitations though. Highly aesthetic or aerodynamic components are largely absent, and cylindrical holes are represented by multiple faces rather than a single cylindrical surface, reflecting the modelling and export conventions of the originating CAD systems.
\subsection{Part Selection}
To obtain individual parts, components were extracted from leaf nodes of the assembly hierarchy. Parts containing multiple solids were excluded, as UV-Net \cite{9577057} cannot process them without modification.
To balance dataset size and geometric diversity, each part was represented by a feature vector $t$ consisting of face-type and edge counts.
Collecting these into a matrix $T = \{t_1, t_2, \dots, t_n\}$, we reduced redundancy by keeping only one representative when multiple parts shared identical statistics, which we define as
\begin{equation}
\tilde{T} = \{ t_i \in T \mid t_i \neq t_j \ \text{for all } j < i \}.
\end{equation}
This ensured coverage across a broad range of geometric characteristics while avoiding redundancy.
%
\paragraph{Labelling policy:}
Annotating holes is non-trivial, since the distinction between hole and non-hole faces is sometimes ambiguous.
For example a torus, contains a geometric hole but is represented as a single B-Rep face. Similar ambiguities occur for holes embedded in blended or curved surfaces. 
Unlike automatically generated labels, human annotation requires resolving ambiguities through functional interpretation.
We therefore adopted a consistent policy: only faces forming cavity surfaces relevant to fastener placement were annotated as hole faces. Simple holes formed by cylindrical surfaces between two planar faces meeting at 90° were always labelled, regardless of fastener use, to ensure consistent coverage of standard geometries. 
Consequently, some geometrically valid holes remain unlabelled and may be counted as false positives during evaluation.
%

%% file: sec/5_experiments.tex
\section{Experiments}
We first define evaluation metrics tailored to the task of detecting fastener-relevant holes, ensuring that performance is measured in a meaningful way. We then present results on generalisation, dataset size, and IoU thresholds, which highlight the strengths and limitations of the methods. All experiments use optimised parameters to enable fair comparison. Parameter optimisation is detailed in a subsequent section.
\subsection{Evaluation Metrics}  
The evaluation metrics are designed to assess the ability to detect the faces of a hole in a meaningful way. For this we evaluate on a per-hole basis, i.e. clusters of faces, as this reflects the ability to detect holes properly. On this level, the evaluation takes the area of the faces into account, as missing a small face is considered less relevant than missing a large one. Let $L$ be the set of faces belonging to a labelled hole, and $F_h$ the set of faces detected for a hole. As the output of the method is not structured as a direct list of holes, $F_h$ is created by clustering adjacent faces using the face adjacency graph. With $A(i)$ denoting the area of face $i$, the Intersection-over-Union is defined as  
\[
\mathrm{IoU}(L,F_h) = \frac{\sum_{i \in L \cap F_h} A(i)}{\sum_{i \in L} A(i) + \sum_{i \in F_h} A(i) - \sum_{i \in L \cap F_h} A(i)}.
\]  
A hole is considered detected if $\mathrm{IoU}(L,F_h) \geq \tau$, with $\tau$ a chosen threshold. Detections above this threshold count as true positives, detections below as false negatives, while clusters of adjacent detected faces with no overlap to a labelled hole are considered false positives.
As UV-Net and AAGNet were originally designed for segmentation, where relevant, an additional set of metrics is reported on a per-face basis. Here, recall and precision values express how well the network segments hole faces in general. These numbers serve as a fair comparison, but hold less weight than the per-hole evaluation, which better reflects the actual task of detecting complete holes.  
\subsection{Generalisation}\label{sec:generalisation}
%
We first evaluate all methods on our benchmark dataset using the optimised parameters from Section~\ref{ssec:optimisation_uvnet}. Results are reported both per-hole and per-face in Table~\ref{tab:opt_results_all_methods}. 
UV-Net and AAGNet achieve the best performance when trained and evaluated on the same dataset (Table~\ref{tab:opt_results_all_methods}), indicating that learning-based methods can effectively model fastener-relevant hole geometry when representative training data is available.
In contrast, the rule-based methods exhibit a recall-precision trade-off, with GWS favouring recall and MVR favouring precision, through its stricter validation strategy.
%
%
\begin{table}[!tbp]
    \footnotesize 
    \setlength{\tabcolsep}{5pt} 
    \centering
    \begin{tabular}{lcccccc}
        \toprule
        & \multicolumn{3}{c}{Per-hole} & \multicolumn{3}{c}{Per-face} \\
        \cmidrule(lr){2-4} \cmidrule(lr){5-7}
        Method & Recall & Prec. & F1 & Recall & Prec. & F1 \\
        \midrule
        AAGNet & 0.944 & 0.810 & 0.871 & 0.912  & 0.932 & 0.922 \\
        UV-Net & 0.934 & 0.896 & 0.914 & 0.883 & 0.933 & 0.920 \\
        GWS    & 0.927 & 0.578 & 0.712 & 0.977 & 0.644 & 0.733 \\
        MVR    & 0.728 & 0.870 & 0.793 & 0.768 & 0.898 & 0.871 \\
        \bottomrule
    \end{tabular}
        \caption{Results with models trained and tested on our dataset, reported per-hole and per-face}
    \label{tab:opt_results_all_methods}
\end{table}
To test generalisation, we evaluated all methods on 200 parts from MFCAD++ \cite{hierarchicalcadnet2022}, ensuring no colliding holes, faulty labels, or missing and extra faces.
Here the relative ranking changes substantially when transferring to MFCAD++ (Table~\ref{tab:generalisation}). Both learning-based methods suffer performance degradation, whereas GWS and MVR remain comparatively stable. However, the effect differs markedly between UV-Net and AAGNet. While UV-Net retains a meaningful level of performance, AAGNet almost completely fails to generalise across datasets. This indicates sensitivity to changes in modelling conventions and dataset characteristics.
%
%
For UV-Net, the largest performance drop occurs for cylindrical holes whose local geometric context differs from the training data. In our benchmark, such holes are frequently represented by multiple faces, whereas they are commonly represented by a single face in MFCAD++.
AAGNet experiences an even larger degradation, suggesting greater reliance on contextual patterns that do not transfer across datasets. MVR mainly fails on long holes when none of the sampled viewpoints observes the hole approximately along its axis. Examples of failure modes are provided in Appendix \ref{app:failure_modes}.
%
\begin{table}[ht]
\footnotesize
\centering
\begin{tabular}{lcccccc}
    \toprule
    & \multicolumn{3}{c}{Per-hole}
    & \multicolumn{3}{c}{Per-face} \\
    \cmidrule(lr){2-4} \cmidrule(lr){5-7}
    Method & Recall & Prec. & F1 & Recall & Prec. & F1 \\
    \midrule
    AAGNet & 0.013 & 0.357 & 0.026 & 0.023 & 0.433 & 0.044 \\
    UV-Net & 0.468 & 0.933 & 0.623 & 0.381 & 0.957 & 0.545 \\
    GWS    & 0.889 & 0.805 & 0.845 & 0.983 & 0.768 & 0.871 \\
    MVR    & 0.617 & 0.987 & 0.760 & 0.577 & 0.987 & 0.729 \\
    \bottomrule
\end{tabular}
\caption{Generalisation study with models trained on our dataset and tested on MFCAD++.}
\label{tab:generalisation}
\end{table}
\subsection{Impact of Dataset size}    
\begin{figure}[!htbp]
    \centering
    \includegraphics[width=\linewidth]{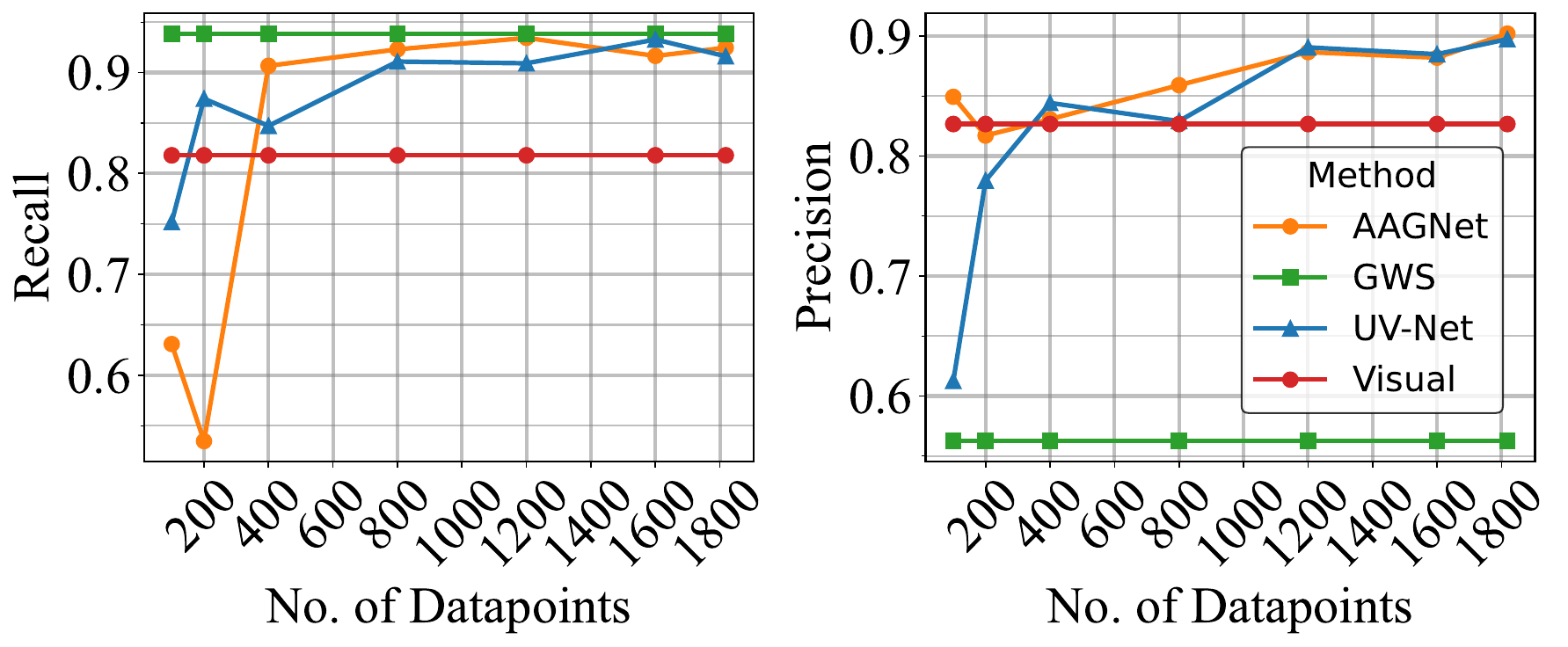}
    \caption{Recall and precision values for all methods at different training dataset sizes}\label{fig:n_datapoint_test}
\end{figure}

The previous section shows that UV-Net struggles to generalise on heterogeneous datasets. This raises the question of how many datapoints are required for UV-Net to perform well.
For this, training and validation subsets of different sizes were created, while the test set remained fixed at 20\% of the dataset. Within each subset, 20\% was used for validation and the remainder for training.
UV-Net and AAGNet performance depends strongly on dataset size (Figure \ref{fig:n_datapoint_test}), whereas the rule-based methods remain constant because they do not require training data.
On small subsets, GWS and MVR outperform the learning-based methods. As more data becomes available, UV-Net and AAGNet improve substantially and eventually surpass the rule-based methods, though instability can still negate gains, as seen in the drop between 1,200 and 1,600 datapoints.
\subsection{IoU Thresholds}
\begin{figure}
\centering
\begin{subfigure}[b]{\linewidth}
    \centering 
    \includegraphics[width=\linewidth]{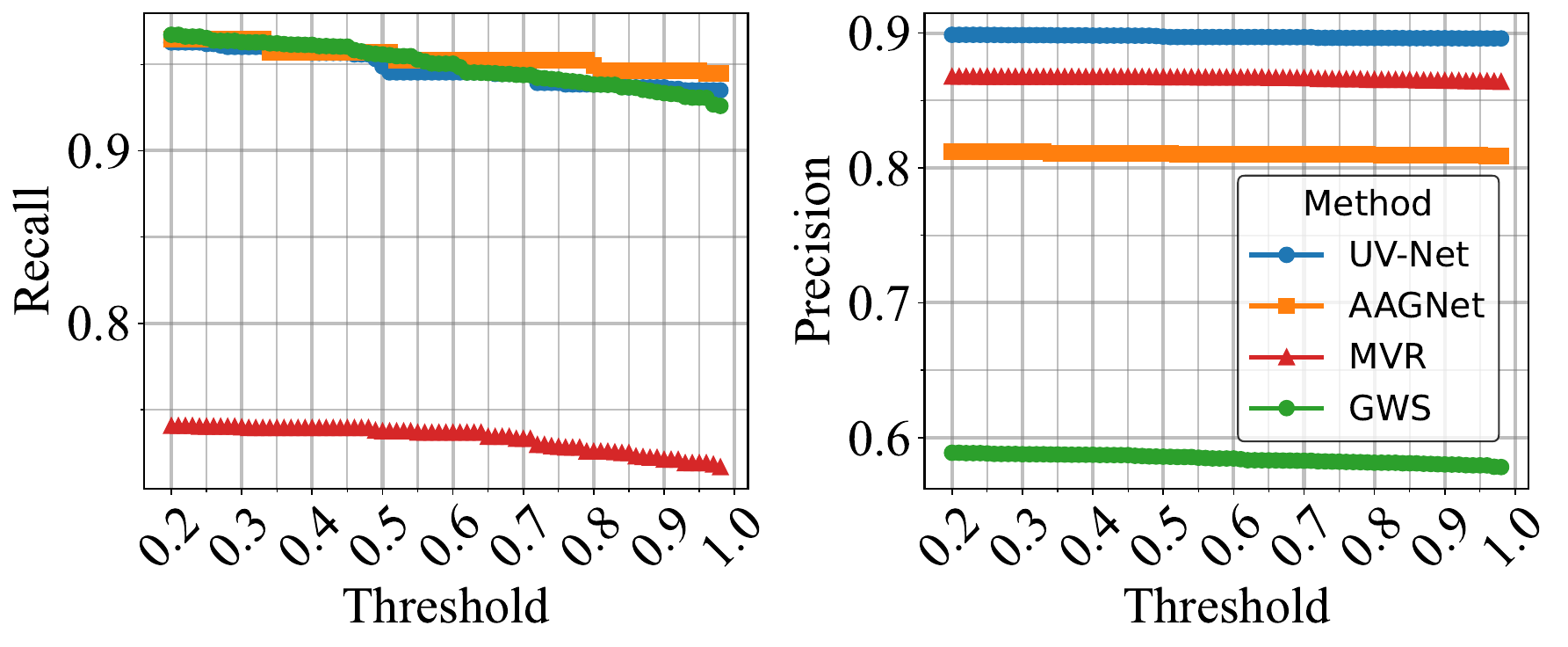} 
\end{subfigure}
\vspace{1mm} 
\begin{subfigure}[b]{\linewidth}
    \centering 
    \includegraphics[width=\linewidth]{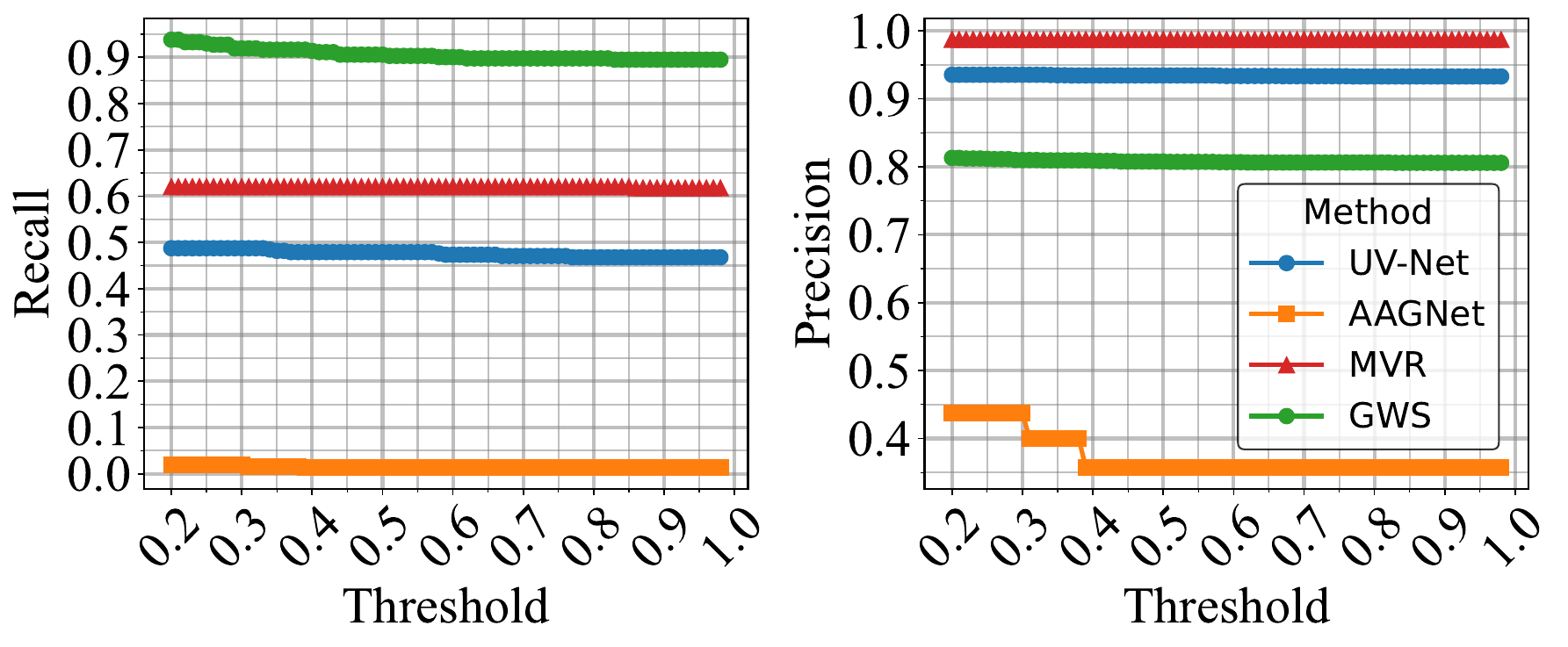}
\end{subfigure} 
\caption{Recall and precision at different IoU thresholds on HDD (top) and MFCAD++ (bottom).} 
\label{fig:iou_thresholds} 
\end{figure}

To evaluate whether the chosen IoU threshold $\tau$ removes a substantial number of faces, and thereby affects the practicality of post-processing partially detected holes, we analyse the effect of varying the threshold on recall and precision.
As shown in Figure~\ref{fig:iou_thresholds}, precision remains largely unaffected by changes in the IoU threshold, while recall decreases only slightly. This indicates that methods tend to either detect a hole almost completely or miss it altogether. UV-Net and AAGNet exhibit similar recall across the entire threshold range, suggesting that both learning-based methods identify a comparable number of holes. However, AAGNet consistently achieves lower precision, indicating a higher false-positive rate. Regarding recall, the learning-based methods performs slightly worse than GWS at low thresholds and slightly better at high thresholds, although these differences lie within the variance observed across optimisation runs and can therefore be regarded as comparable. The limited decline in recall for the rule-based methods is likely explained by the validation steps that ensure only valid hole topologies are retained.    
%
When applying the same threshold study to the MFCAD++ subset (Section~\ref{sec:generalisation}), the trends remain similar. UV-Net, GWS, and MVR again exhibit only minor changes in recall and precision across the threshold range. The recall of AAGNet also remains relatively stable, albeit at a substantially lower level due to its failure to generalise across datasets. AAGNet achieves higher precision at very low thresholds, indicating that some of its few successful detections correspond only to partial overlaps with the annotated holes.
These results indicate that high IoU thresholds are feasible, as they provide cleaner detections with only minor losses in recall and precision, thereby reducing the need for post-processing.
%
%
\subsection{Parameter optimisation}
\begin{figure}[!tbp]
    \begin{subfigure}{0.48\textwidth}
        \centering
        \includegraphics[width=\columnwidth]{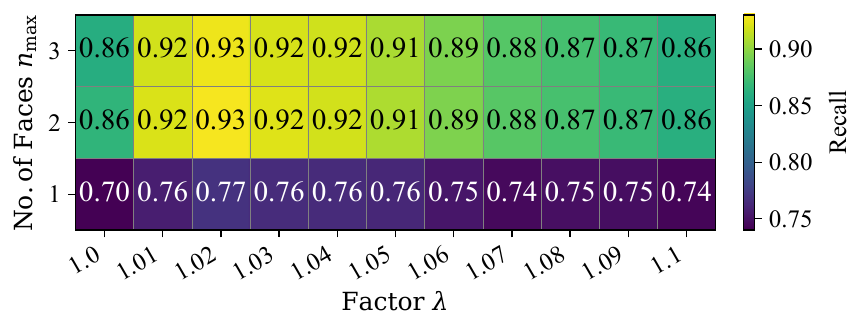}
    \end{subfigure}
    \vfill
    \begin{subfigure}{0.47\textwidth}
        \centering
        \includegraphics[width=\columnwidth]{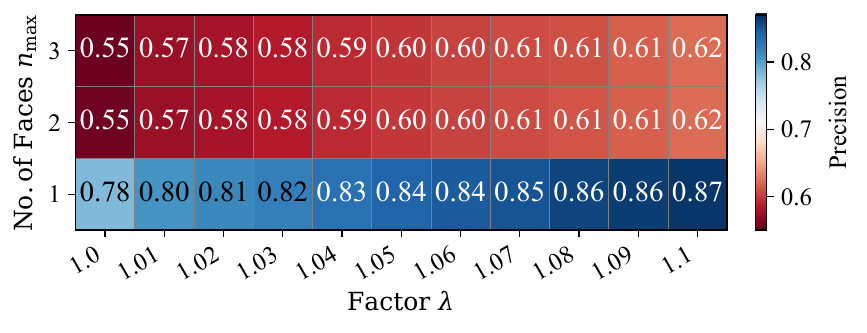}
    \end{subfigure}
    \caption{Parameter optimisation results for the graph-wire-search method}
    \label{fig:heatmap_struct_opt}
\end{figure}

\begin{figure}[h]
    \centering
    \begin{subfigure}[b]{0.495\linewidth}
        \centering
        \includegraphics[width=\linewidth]{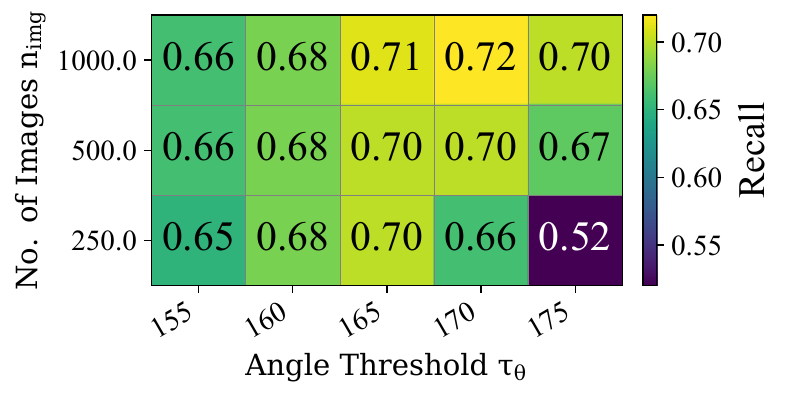}
    \end{subfigure}
    \hfill
    \begin{subfigure}[b]{0.495\linewidth}
        \centering
        \includegraphics[width=\linewidth]{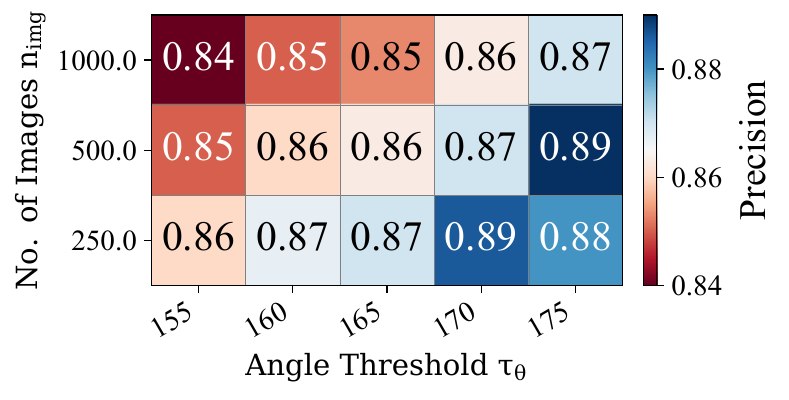}

    \end{subfigure}
    \caption{Parameter optimisation results of the Multiview Rendering method}

    \label{fig:cv_opt_results}
\end{figure}

\begin{figure} 
    \centering 
    \begin{subfigure}[b]{0.495\linewidth} 
        \centering
        \includegraphics[width=\linewidth]{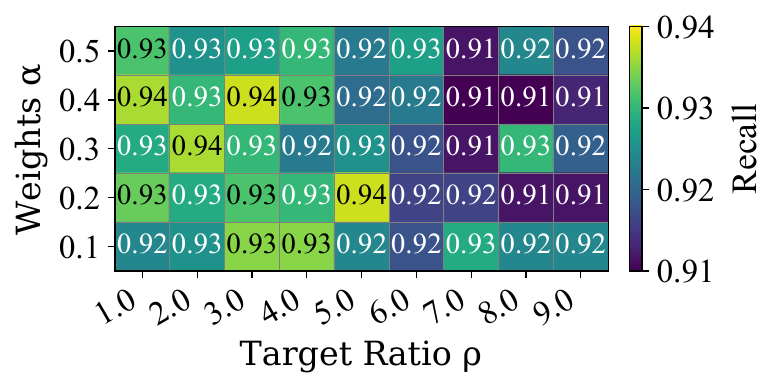} 
    \end{subfigure}
    \hfill 
    \begin{subfigure}[b]{0.495\linewidth}
        \centering 
        \includegraphics[width=\linewidth]{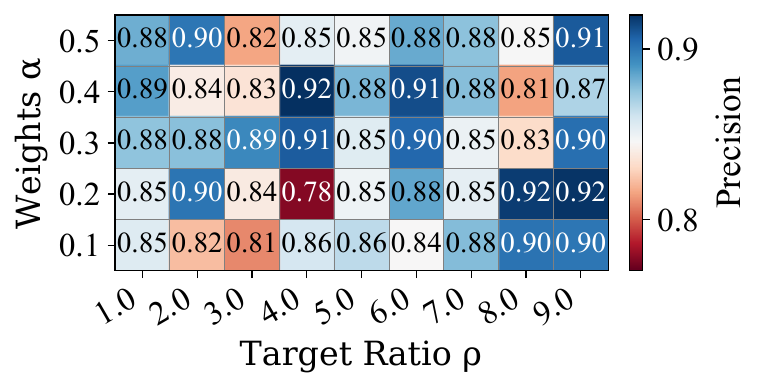} 
    \end{subfigure} 
    \vspace{1mm} 
    \begin{subfigure}[b]{0.495\linewidth}
        \centering
        \includegraphics[width=\linewidth]{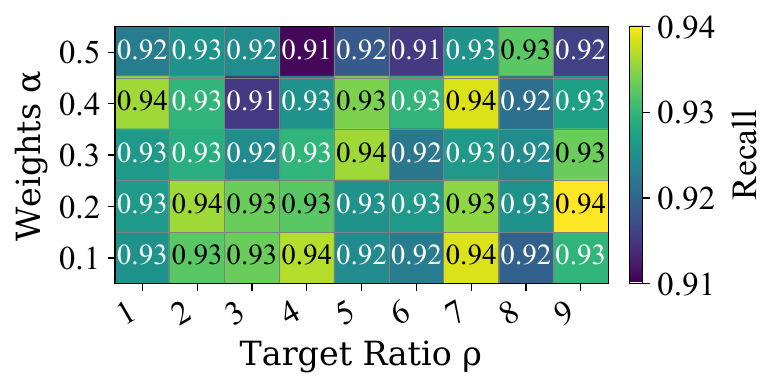} 
    \end{subfigure}
    \hfill
    \begin{subfigure}[b]{0.495\linewidth}
        \centering
        \includegraphics[width=\linewidth]{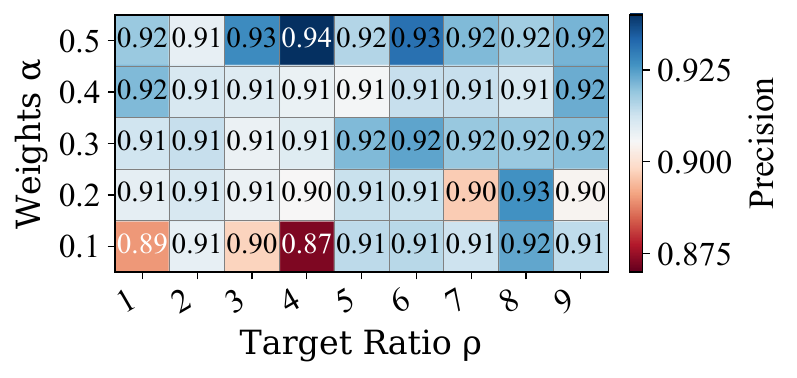} 
    \end{subfigure} 
    \caption{Recall and precision for AAGNet (top) and UV-Net (bottom) under different class weights and oversampling ratios.} 
    \label{fig:learning_opt_results}
\end{figure}

A grid search was performed to determine suitable parameters for each approach. For per-hole evaluation, a threshold of 0.99 was applied to the IoU, requiring holes to be detected nearly perfectly. This corresponds to a fully automated setting without human post-processing.
\paragraph{Graph-Wire-Search}
For GWS, both the maximum number of faces $n_{\mathcal{F}^*,\mathrm{max}}$ used for wire generation and the hollowness factor $\lambda$ were varied. The number of faces ranged from 1 to 3, while $\lambda$ was varied from 1.00 to 1.10 in steps of 0.01.
The results in Figure~\ref{fig:heatmap_struct_opt} indicate that GWS is relatively insensitive to parameter choice. Using two or three faces produces nearly identical results, suggesting that the method remains stable for more complex hole boundaries. A value of $\lambda=1.02$ provides the best trade-off between recall and precision by rejecting low-curvature regions that are unlikely to belong to holes. Larger values further improve precision but at the cost of reduced recall.
\paragraph{Multiview Rendering}
For the visual approach, both the angle $\theta^\circ$ and the number of generated images $n_\mathrm{img}$ were varied, with results shown in Figure~\ref{fig:cv_opt_results}. Performance improves with increasing numbers of rendered views, confirming that coverage is a key factor for reliable hole detection. The influence of $\theta$ is less pronounced between $155^\circ$ and $165^\circ$, where the relaxed matching criterion reduces sensitivity to the exact threshold. Larger angular tolerances, however, also admit more infeasible candidate faces, reducing recall. Since image generation dominates runtime, the number of rendered views provides a practical trade-off between computational cost and performance. The highest recall is achieved at $\theta = 170^\circ$ and $n_\mathrm{img}=1000$, while maintaining high precision. These settings were therefore used for all subsequent experiments.
\paragraph{UV-Net \& AAGNet}\label{ssec:optimisation_uvnet}  
For both learning-based methods, the original training configurations were retained where possible. As our benchmark exhibits a strong class imbalance between hole and non-hole faces, we investigated class weighting and oversampling. The hole-class weight $\alpha$ was varied between 0.5 and 0.9 in steps of 0.1, while oversampling was used to achieve target ratios of $1:\rho$ with $\rho \in [1,9]$. The same optimisation strategy was applied to UV-Net and AAGNet.
The results shown in Figures \ref{fig:learning_opt_results} are averaged over three runs. For both models, no clear trend emerges with respect to class weighting or oversampling. The variation between repeated runs is often comparable to the difference across parameter settings, suggesting limited sensitivity to these hyperparameters. Consequently, the weight and target ratio were selected independently based on their average recall and precision across runs and across all values of the respective other parameter. This resulted in a target ratio of 7 and a hole-class weight of 0.2 for UV-Net, and a target ratio of 3 and a weight of 0.1 for AAGNet.
%
%

%

%% file: sec/6_conclusion.tex
\section{Conclusion}
We presented a comparative study of rule-based and learning-based methods for fastener-relevant hole detection in CAD models. To this end, we introduced two rule-based approaches and a manually annotated benchmark derived from engineering CAD assemblies. 
The graph-wire-search method achieves high recall, while the multiview rendering method favours precision. Both methods generalise robustly across datasets without requiring training data. 
In contrast, UV-Net and AAGNet achieve stronger performance when trained and evaluated on the same dataset, but require sufficient annotated data and exhibit reduced robustness under domain shift. 
These results highlight a fundamental trade-off between learning-based and rule-based approaches. While learning-based methods benefit from representative annotations and sufficient training data, rule-based methods currently provide the most reliable performance across heterogeneous CAD datasets. 



%% file: sec/7_appendix.tex
\clearpage
\appendix

\section{Industrial Assembly Dataset statistics}\label{sec:dataset_stats}
    Table \ref{tab:dataset_stats} summarises the main characteristics of the dataset, which are relevant for understanding the challenges in model training and evaluation. The dataset comprises 2202 models with a total of 197003 faces. Among these, 14471 are annotated as hole faces, corresponding to 7.3\% of the dataset, thereby showing a strong class imbalance.
    
    \begin{table}[H]
        \centering
        \small
        \begin{tabular}{l r}
            \toprule
            Statistic & Count / Ratio \\
            \midrule
            Hole faces             & 14{,}471 \\
            Models                 & 2{,}202 \\
            Total faces            & 197{,}003 \\
            Hole face proportion   & 0.073 \\
            Holes                  & 6{,}864 \\
            Avg.\ holes per model  & 3.02 \\
            \bottomrule
        \end{tabular}
        \caption{Dataset statistics}
        \label{tab:dataset_stats}
    \end{table}

\section{Derivation of Camera Distance}
\label{app:camera_distance}

To determine the optimal camera distance, the object must fill the field of view while remaining fully visible from all angles. Let $H$ denote the sensor height (equal to its width for symmetry), $f$ the focal length, and $\theta$ the angular field of view. Following \cite[Ch.~2]{Szeliski2010}, the relation between $\theta$, $H$, and $f$ is given by

\begin{equation}
    \tan\!\left(\tfrac{\theta}{2}\right) = \frac{H}{2f}.
\end{equation}

For a given working distance $WD$, the linear field of view $LFOV$ can be expressed as
\begin{equation}
    LFOV = 2 \cdot WD \cdot \tan\!\left(\tfrac{\theta}{2}\right).
\end{equation}

Rearranging gives
\begin{equation}
    f = \frac{H \cdot WD}{LFOV}.
\end{equation}

To guarantee full object visibility, $LFOV$ is set equal to the object’s maximum diagonal, which we approximate by the bounding box diagonal $d_{\mathrm{BB}}$,
\begin{equation}
    f = \frac{H \cdot WD}{d_{\mathrm{BB}}}.
\end{equation}

Solving for the working distance yields
\begin{equation}
    WD = \frac{f \cdot d_{\mathrm{BB}}}{H}.
\end{equation}

\section{Qualitative Examples of Failure Modes}\label{app:failure_modes}
\begin{figure}[hb]
  \begin{subfigure}[b]{0.55\linewidth}
    \centering
    \includegraphics[width=\linewidth]{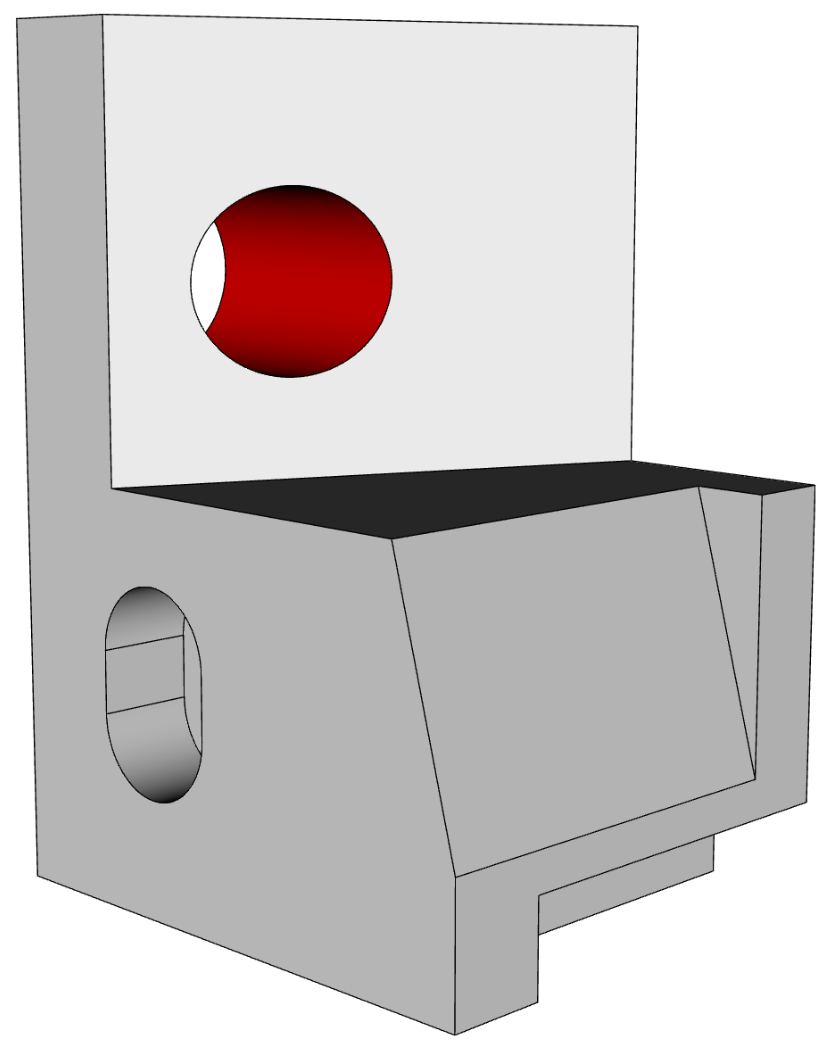} 
    \caption{Hole missed by both learning-based approaches, illustrating a representative failure case.}
    \label{fig:learning_failure}
  \end{subfigure}%
  \hfill
  \begin{subfigure}[b]{0.38\linewidth}
    \centering
    \includegraphics[width=\linewidth]{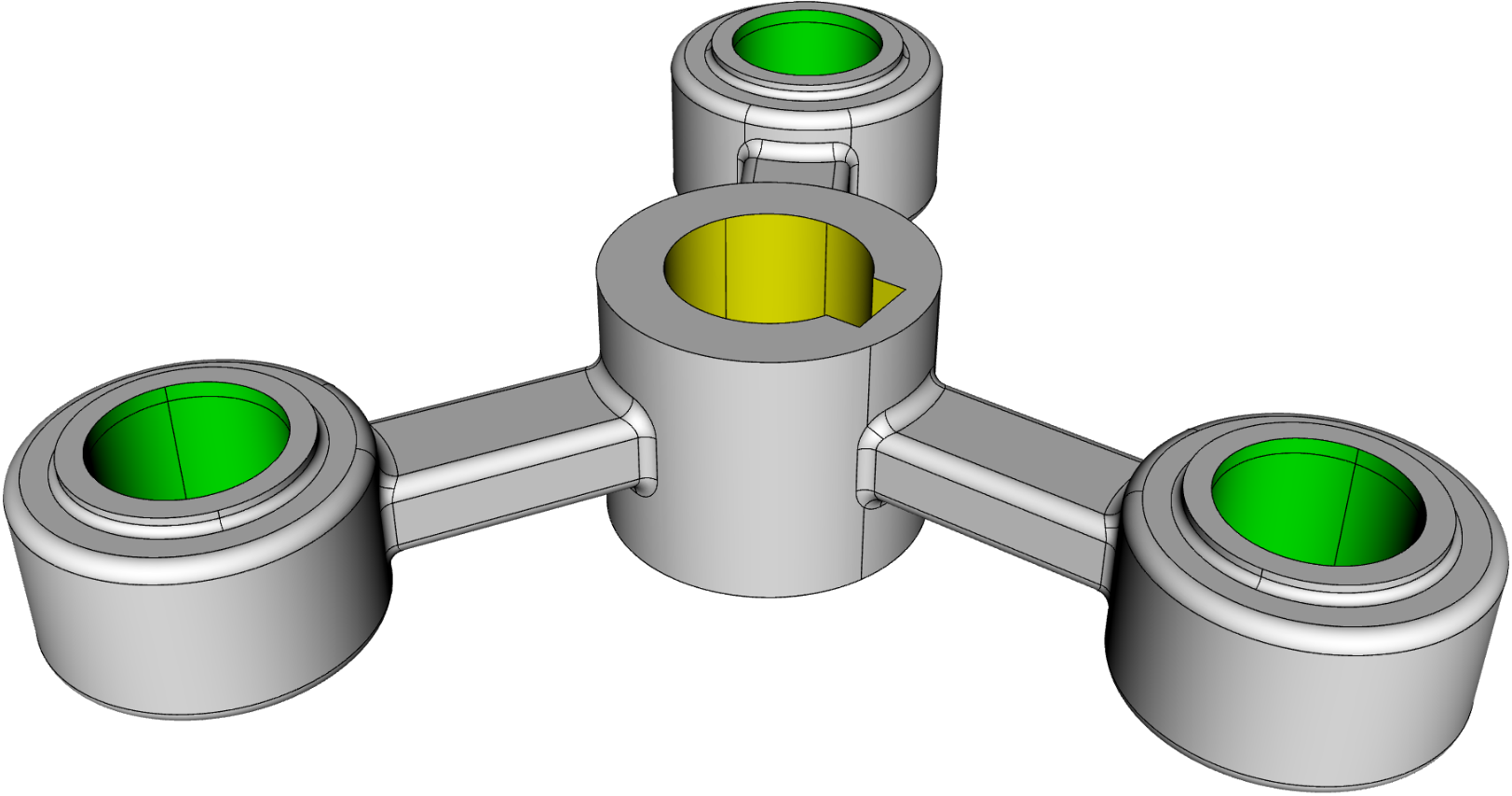} \caption{GWS detects additional geometrically valid but unlabelled holes.}
    \label{fig:gws_unlabelled_holes} 
    \vspace*{\baselineskip} 
    \includegraphics[width=\linewidth]{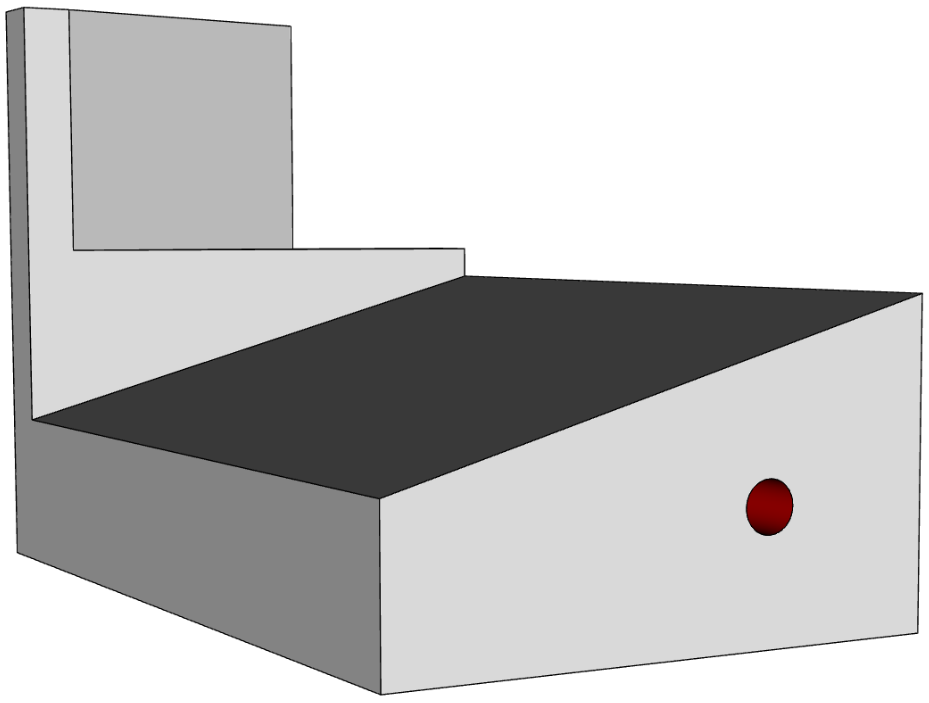}
    \caption{MVR fails on a deep hole due to insufficient view alignment.}
    \label{fig:mvr_long_hole} 
  \end{subfigure}
  \caption{Representative failure modes. Green indicates correctly detected holes, red indicates missed holes, and yellow indicates detected but unlabelled holes.} 
    \label{fig:qualitative_examples}
\end{figure}

To complement the quantitative evaluation, Figure \ref{fig:qualitative_examples} shows representative failure cases. Green indicates correctly detected holes, red indicates missed holes, and yellow indicates detected but unlabelled holes. The examples illustrate common failure modes of the considered methods and provide additional insight into the results discussed in Section~\ref{sec:generalisation}.

